\documentclass[runningheads]{llncs}

\usepackage{eccv}

\usepackage{eccvabbrv}

\usepackage{graphicx}
\usepackage{booktabs}

\usepackage[accsupp]{axessibility}  

\usepackage{hyperref}

\usepackage{orcidlink}

\usepackage{xspace}
\usepackage{enumitem}
\usepackage{wrapfig}
\usepackage{booktabs,multirow,adjustbox,tabularray}
\usepackage{mathtools,nicefrac}
\usepackage{makecell}
\usepackage{capt-of}
\usepackage[ruled,vlined]{algorithm2e}
\SetAlgoNlRelativeSize{0} % Set to zero to remove semicolons

\usepackage{etoolbox}
\providetoggle{beginFlag}
\settoggle{beginFlag}{true}

\SetAlgoSkip{AlgoSkipBeforeAndAfter}

\usepackage[dvipsnames]{xcolor}
\newcommand{\silenced}[1]{}

\newcommand{\ourmethod}{SplatPainter\xspace}

\usepackage{dsfont}

\newif\ifshowedits
\newcommand{\addeditor}[3]{%
  \definecolor{#1color}{rgb}{#3}
  \expandafter\newcommand\csname #1\endcsname[1]{%
  \ifshowedits
    {\color{#1color} ##1}%
  \else
    {##1}%
  \fi
  }%
  \expandafter\newcommand\csname #1rmk\endcsname[1]{%
  \ifshowedits
    {\color{#1color} {\bf [#2: ##1]}}
  \fi
  }%
  \expandafter\newcommand\csname #1rpl\endcsname[2]{%
  \ifshowedits
    {{\color{#1color} ##1} \sout{##2}}
  \else
    {##1}
  \fi
  }%
}

\definecolor{darkgreen}{RGB}{0,110,0}
\definecolor{darkred}{RGB}{170,0,0}

\addeditor{yifan}{YW}{0.7, 0.0, 0.7}
\addeditor{yang}{YZ}{0.0, 0.0, 0.8}
\addeditor{gordon}{GW}{0.1, 0.9, 0.2}

\showeditstrue
\begin{document}

\title{SplatPainter: Interactive Authoring of 3D Gaussians from 2D Edits via Test-Time Training}

\titlerunning{SplatPainter}

\author{Yang Zheng\textsuperscript{1,2} 
\quad Hao Tan\textsuperscript{2} \quad Kai Zhang\textsuperscript{2} \quad Peng Wang\textsuperscript{2} \\ \quad Leonidas Guibas\textsuperscript{1}  \quad Gordon Wetzstein\textsuperscript{1} \quad Wang Yifan\textsuperscript{2} 
}

\authorrunning{Zheng et al.}

\institute{Stanford University \and
Adobe }

\maketitle
\begin{center}
% \vspace{-10pt}
\includegraphics[width=\textwidth]{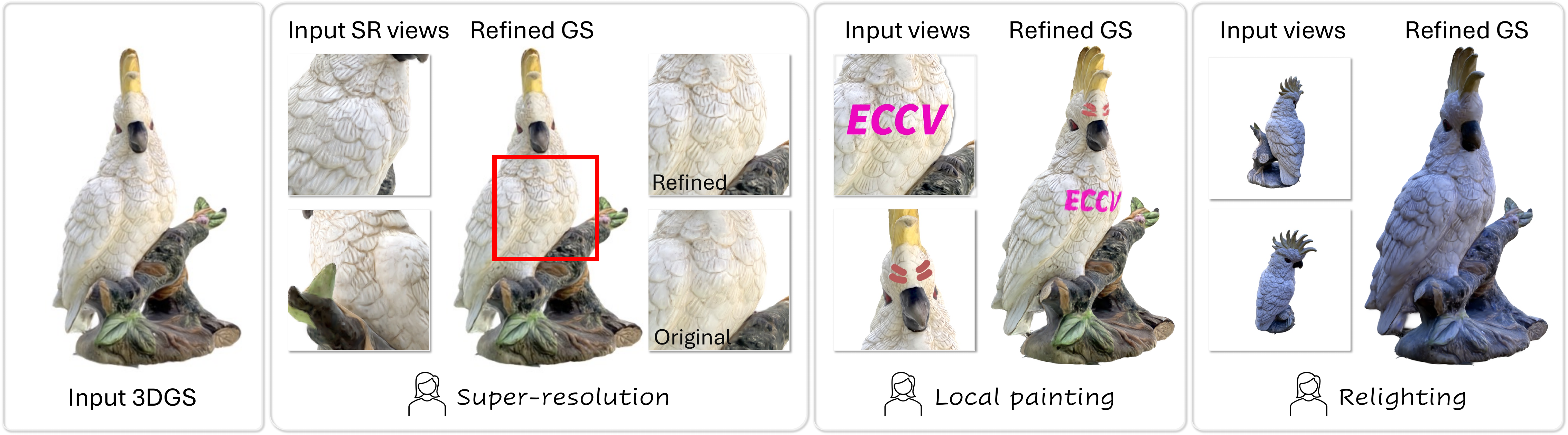}
  \captionof{figure}{We present \ourmethod, a feedforward method to support interactive and continuous authorship of 3D Gaussian assets through intuitive 2D edits, \eg, high-fidelity local detail refinement, local paint-over, and consistent global recoloring, while maintaining 3D consistency and the original texture and structure details. }
  \label{fig:teaser}
\end{center}

\begin{abstract}
The rise of 3D Gaussian Splatting has revolutionized photorealistic 3D asset creation, yet a critical gap remains for their interactive appearance refinement and editing. Existing approaches based on diffusion or optimization are ill-suited for this task, as they are often prohibitively slow, lack the precision for fine-grained control, or—most importantly—are fundamentally destructive to the original asset's identity. To address this, we introduce SplatPainter, a state-aware feedforward model that enables continuous, high-fidelity appearance editing of 3D Gaussian assets from user-provided 2D views. Our method directly predicts updates to the attributes of a compact, feature-rich Gaussian representation and leverages Test-Time Training to create a state-aware, iterative workflow. The versatility of our approach allows a single architecture to perform diverse tasks, including high-frequency local detail refinement, local paint-over, and consistent global recoloring, all at interactive speeds while strictly preserving the asset’s original structure. Our project website is at: \url{https://y-zheng18.github.io/SplatPainter/}.
\end{abstract}    
\section{Introduction}
\label{sec:intro}

High-quality 3D content creation is fundamental to a growing number of industries, from gaming and virtual reality to e-commerce and digital twins. Recently, 3D Gaussian Splatting (3DGS)~\cite{kerbl3Dgaussians} has emerged as a leading representation for its real-time photorealistic rendering and its ability to capture complex details like hair and fur more effectively than traditional meshes. As text- and image-to-3D pipelines~\cite{wang2025diffusion,li2024advances} built on GS become more powerful, they enable the rapid creation of assets. However, these one-shot generation methods typically represent a ``starting point'' rather than a final product; their focus on global coherence often comes at the cost of local detail, creating an urgent need for efficient workflows to refine the appearance of these assets post-creation.

\begin{figure}[t!]
    \centering
    \includegraphics[width=\textwidth]{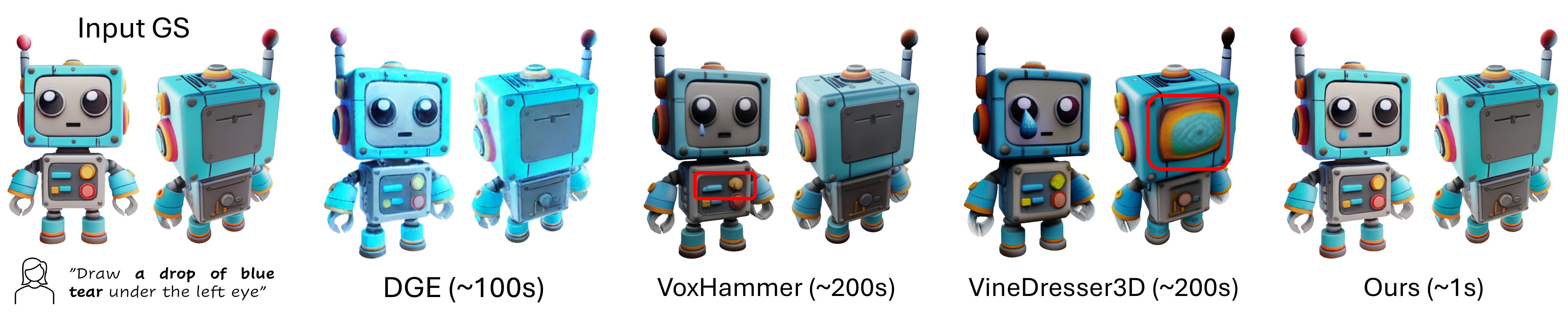}
  \captionof{figure}{\textbf{Challenges in Controllability and Identity Preservation}. Existing methods struggle to propagate precise 2D edits into 3D space. Approaches that rely on multiview generation (e.g., DGE~\cite{chen2024dge}) struggle to match user intention (e.g., the drop of tear) and often result in 3D inconsistencies and blurry textures. Even direct 3D methods (e.g., VoxHammer~\cite{li2025voxhammer}, VineDresser3D~\cite{chi2026vinedresser3d}) can introduce global identity drift during localized updates (e.g., color shift and detail distortion). In contrast, \ourmethod provides precise, intent-matching updates—leveraging external editors like Nano Banana for high-quality guidance—while strictly maintaining the original asset's identity.}
  \label{fig:teaser2}
\end{figure}

The critical missing piece in the modern 3D workflow is a framework for interactively and intuitively editing the appearance of existing 3DGS assets. For content creation to be truly powerful, users need to make precise adjustments with familiar 2D tools, applying edits to arbitrary views—including close-ups for fine-grained refinement. This is a non-trivial challenge that requires addressing three core difficulties: (1) achieving precise, controllable updates that faithfully follow user intent; (2) preserving the original identity and intricate structural details of unedited regions; and (3) operating at interactive speeds. As illustrated in \cref{fig:teaser2}, maintaining identity is not a "given" even for texture-only edits. Existing tools often introduce significant and unintended identity drift or fail to match the user's specific 2D intent, leading to results that are either perceptually inconsistent with the original asset.

Current 3D editing approaches are fundamentally ill-suited for this task of interactive refinement.
Many methods rely on an intermediate step of generating multiview images to match a 2D edit before lifting them to 3D via reconstruction~\cite{barda2025instant3dit, li2025cmd, chi2025disco3d, chen20243d, bar2025editp23, erkocc2025preditor3d, qi2024tailor3d, gaussctrl2024, chen2024gaussianeditor}. This process is prone to 3D inconsistencies that destroy original textural details and lead to "blurry" results that fail to capture the user's specific intent.
Alternatively, methods that employ diffusion directly on structured 3D latents~\cite{li2025voxhammer, chi2026vinedresser3d} bypass multiview generation but suffer from ``spatial edit leakage''. In these cases, appearance changes frequently bleed into unintended regions -- a limitation common to latent diffusion models~\cite{}.
Regardless of the underlying paradigm, these methods share a prohibitive latency. Whether based on iterative diffusion sampling or per-scene optimization, they are far too slow for real-time feedback, often requiring hundreds of seconds for a single change and breaking the fluid creative process required by artists.

In this work, we introduce \ourmethod, a state-aware feedforward framework designed for fast and identity-preserving appearance editing of 3D Gaussian assets. By resolving 2D-to-3D consistency directly in the Gaussian representation, we bypass the need to finetune or repurpose specific generative models. Instead, our approach acts as a versatile appearance-lifting module that can incorporate 2D edits from diverse sources—including manual user paint-overs, super-resolution models, or advanced generative editors like Nano Banana~\cite{nanobanana2026} -- without requiring the destructive multiview regeneration typical of existing workflows. Our model directly predicts updates to Gaussian attributes (e.g., color and opacity) from these image-space edits, ensuring the asset’s original structural integrity remains intact. To facilitate a continuous creative process, we utilize Test-Time Training (TTT)~\cite{sun2020test,zhang2025test,wang2025test} to make the model state-aware. This allows the network to adapt to user inputs on the fly, integrating new appearance information while maintaining the underlying identity of the original 3D asset.

Our main contributions can be summarized as follows:
\begin{enumerate}[nosep]
\item \textbf{A Novel Feedforward Editing Framework:} We introduce \ourmethod, a state-aware feedforward model for the interactive and continuous editing of 3DGS. By leveraging TTT, our method iteratively incorporates user edits from 2D views, directly updating the attributes of a compact and feature-rich Gaussian representation without destructive regeneration.
\item \textbf{Versatile and Intuitive Applications:} We evaluate our framework on two novel and practical applications: high-fidelity local detail refinement and consistent global relighting. The unified approach provides users with precise control over both fine-grained texture details and broad appearance changes from 2D edit(s).
\item \textbf{State-of-the-Art Performance at Interactive Speeds:} Through extensive experiments, we show that \ourmethod significantly outperforms existing methods in both quality and speed. Our approach achieves superior visual results while operating at interactive rates, making it a truly practical solution for unlocking fluid, real-time creative workflows in 3D content authoring.
\end{enumerate}

\section{Related Work}
\label{sec:related}
Our work lies at the intersection of editing for modern 3D representations like NeRFs~\cite{mildenhall2021nerf} and 3D Gaussian Splatting (3DGS)~\cite{kerbl3Dgaussians}, and continuous 3D reconstruction. We begin by reviewing the evolving ecosystem of direct GS editing before discussing broader paradigms for image-driven 3D editing and the architectural principles that enable our iterative approach.

\noindent\textbf{The Evolving 3D Gaussian Splatting Ecosystem.}
The wide adoption of 3DGS~\cite{kerbl3Dgaussians} and its plethora of extensions~\cite{fei20243d} has propelled research into making GS assets directly editable, leading to the emergence of interactive tools for manual selection and editing~\cite{SuperSplat, LumaAI, KIRIEngine, Michel2023ProjectNewDepth}. More advanced methods achieve real-time geometric deformation through auxiliary cage structures~\cite{huang2024gsdeformerdirectrealtimeextensible, tong2025cage, xie2024sketch} or use feed-forward networks for tasks like splat densification~\cite{chen2024splatformer, nam2025generative}. 
Other works augment the GS representation itself to enable new modalities like physics simulation~\cite{xie2024physgaussian}. 
While these approaches showcase a clear demand for direct and efficient GS manipulation, they primarily focus on geometry modification. They do not provide a mechanism for our task: intuitive, fine-grained appearance editing guided by 2D image inputs.

\noindent\textbf{Image-Driven 3D Editing.}
Image-driven 3D editing is a central challenge, with three dominant strategies emerging.
The first, an optimization-heavy approach, updates a 3D representation by optimizing it to match a set of edited 2D views. This paradigm is used for tasks like geometry and texture refinement~\cite{ryu2025elevating, chen2024text}, super-resolution~\cite{xia2025close, wan2025s2gaussian, yu2024gaussiansr, feng2024srgs, shen2024supergaussian, xie2024supergs, lee2024disr, yoon2023cross, xia2025enhancing}, and CLIP-guided stylization~\cite{radford2021learning, michel2022text2mesh, wang2022clip, wang2023nerf, song2023blending, wang2023inpaintnerf360, kobayashi2022decomposing, chen2022tango, gao2023textdeformer, hong2022avatarclip, zhou2024feature}. Score Distillation Sampling~\cite{poole2022dreamfusion} is a popular variant, where optimization is guided by text prompts instead of explicit images\cite{chen2024shap, park2023ed, zhang2023text, zhuang2023dreameditor, kamata2023instruct, decatur20243d, dong2024interactive3d, dinh2025geometry, liu2024make, zhuang2024tip, voleti2024sv3d, xu2024tiger, sella2023vox}. Although these methods can produce high-quality results, their reliance on slow, per-scene optimization makes them unsuitable for interactive workflows.
The second strategy, generative reconstruction, uses a multi-view diffusion model to generate consistent 2D edits from a prompt~\cite{barda2025instant3dit, li2025cmd, chi2025disco3d, chen20243d, bar2025editp23, erkocc2025preditor3d, qi2024tailor3d, gaussctrl2024, chen2024gaussianeditor, wen2025intergsedit, haque2023instruct, igs2gs, chen2024dge}, which are then lifted to a new 3D asset using a feedforward reconstruction model~\cite{gslrm2024,hong2023lrm,xu2024grm}. This workflow, while faster than full optimization, is still too slow for real-time use and inherently destructive as it replaces the original asset.
Finally, recent approaches explore direct 3D editing via structured diffusion latents to bypass multiview generation~\cite{li2025voxhammer, xiang2025structured, xiang2025native, chi2026vinedresser3d}. While methods enable more coherent updates by operating in a native 3D latent space, they often suffer from spatial leakage where edits bleed into unedited regions due to the global receptive fields of diffusion priors. Our method, in contrast to these strategies, is fast, non-destructive, and handles fine-grained image edits with strict locality.

\noindent\textbf{Continuous and State-Aware Architectures.}
Our architecture is inspired by continuous reconstruction, where a 3D representation is progressively improved with new observations. This concept spans classical SLAM~\cite{engel2014lsd, mur2015orb, forster2016svo, teed2021droid, zhu2024nicer} and Structure-from-Motion~\cite{agarwal2011building, schonberger2016structure} to modern learning-based methods using recurrent or memory-augmented networks to maintain state~\cite{choy20163d, kar2017learning, sun2021neuralrecon, zhang2022nerfusion, schuster1997bidirectional, graves2012long, jaegle2021perceiver}. Recent frameworks explicitly maintain latent states for continuous updates~\cite{wang20243d, wang2025continuous, chen2025ttt3r}, while in the 3DGS domain, this is often done via continual optimization~\cite{zeng2025gaussianupdate}. Drawing from these state-aware designs, we use test-time training (TTT)~\cite{zhang2025test} to enable our feedforward model to iteratively incorporate user edits, achieving an efficient, online refinement workflow.

\begin{figure}[t!]
  \centering
  \includegraphics[width=\textwidth]{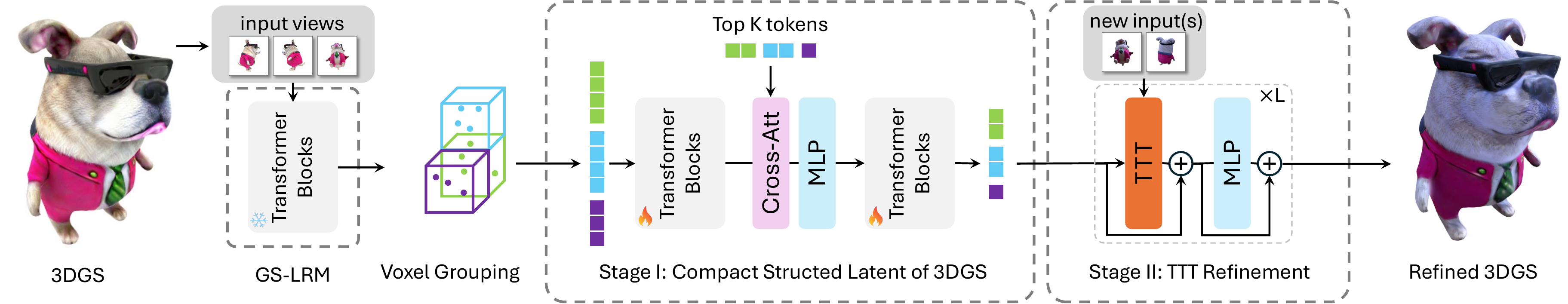}
  \caption{\textbf{Overview.} Given an input 3DGS asset, the framework first performs a one-time preprocessing step. The asset is rendered from multiple views to generate feature-rich inputs from a Gaussian LRM. Stage I compresses this into a compact latent representation via a local transformer. The interactive editing loop in Stage II then iteratively refines this latent representation using new 2D user edits (new input(s)) and a Test-Time Training (TTT) module to produce the final edited 3DGS.}
  \label{fig:pipeline}
  \vspace{-10pt}
\end{figure}
\section{Method}
\label{sec:method}

Given a 3DGS asset $\mathcal{G} = \{g_i\}_{i=1}^N$, our goal is to efficiently update Gaussian attributes—position, scale, opacity, rotation, and spherical harmonics—based on user-provided image-space edits such as zoom-in super-resolution, graffiti-style markings, or global recoloring (see \cref{fig:teaser}). Our key contribution is a feedforward, state-aware pipeline composed of two key components: (1) a compact and feature-rich Gaussian representation extracted via a one-time preprocessing step, and (2) a refinement module with Test-Time Training (TTT) layers~\cite{zhang2025test} that incorporate new user inputs to iteratively update the scene (Fig.~\ref{fig:pipeline}).

\subsection{Compact and Feature-Rich 3DGS}\label{sec:method-1}
Intelligently propagating a 2D edit requires a 3D representation with rich global context. Our strategy is to leverage a Gaussian large reconstruction model (LRM)~\cite{xu2024grm, gslrm2024, chen2024mvsplat} to create a high-fidelity, identity-preserving copy of the asset that is inherently enriched with such features. To accomplish this, we first render the input 3DGS from a set of canonical viewpoints. These rendered images serve as the input for the LRM, which then processes them to generate a new, dense cloud of Gaussians where each point is associated with a rich feature vector reflecting the global context. 
While this provides the necessary context, the LRM's native output is a dense, unstructured cloud of Gaussians—unsuitable for interactive editing due to high redundancy and memory overhead. Therefore, we introduce a one-time preprocessing step: a voxel-based transformer that compresses and regularizes this dense output. This process extracts the globally-informed features into a compact, structured representation ideal for efficient, iterative refinement.

\vspace{1mm}\noindent\textbf{Preliminary: Gaussian LRM.}
Recent Gaussian LRMs provide a strong foundation for generating high-quality Gaussian Splatting assets from sparse multi-view images. A common approach in these frameworks is to process input multiview images $\{I_v\}_{v=1}^V$ with a transformer-based backbone. This backbone typically extract image tokens from patchified multiview images and then feed the tokens through a chain of self-attention transformer blocks to obtain feature tokens $\{F_v\}_{v=1}^n$. These tokens are then decoded to predict per-pixel Gaussian attributes $g_i$. Our work builds on this architectural paradigm to acquire the initial feature-rich Gaussians before our compression stage.

\vspace{1mm}\noindent\textbf{Voxelization.}
To impose a 3D structure, we first reorganize the reconstructed Gaussians into a regular latent grid. We voxelize the scene, grouping nearby Gaussians into local cells. Each resulting voxel $\mathcal{V} = \{(f_k, g_k)\}_{k=1}^{N_k}$ contains $N_k$ Gaussians, where each Gaussian is associated with its feature token $f_k$ from the LRM backbone and its Gaussians attributes $g_k$.

\vspace{1mm}\noindent\textbf{Local Voxel Transformer.}
On top of this structured representation, we introduce a local voxel transformer to produce compact and context-aware latent representations for Gaussians in each voxel. 
% \yifan{wait, this sentence sounds like there is one latent for every voxel, but we have a latent for every gs in the voxel right?}. 
This is a two-stage process of aggregation and distillation. Given a voxel $\mathcal{V} = \{(f_k, g_k)\}_{k=1}^{N_k}$, we first aggregate intra-voxel relations via a transformer encoder consisting of $L_{enc}$ layers of self-attention~\cite{vaswani2017attention}:
\begin{equation}
\mathcal{Z} = \mathrm{TransformerEncoder}(\mathcal{V}),
\end{equation}
where $\mathcal{Z}= \{z_k\}_{k=1}^{N_k}$ is the set of context-aware latent codes for each Gaussian within the voxel. To reduce redundancy and summarize this information, we select the top-$K$ most significant Gaussians based on their opacity $\alpha$ and use their latent codes as queries in a cross-attention mechanism:
\begin{equation}
\tilde{\mathcal{Z}} = \mathrm{CrossAttn}\big(\{z_i\}_{i \in \text{TopK}_\alpha}, \mathcal{Z}\big).
\end{equation}
This step effectively distills the information from all Gaussians in the voxel into a compact set of features. Finally, a transformer decoder with $L_{dec}$ self-attention layers refines these compressed features into the final, compact voxel latent:
\begin{equation}
\tilde{\mathcal{V}} = \mathrm{TransformerDecoder}(\tilde{\mathcal{Z}}).
\end{equation}
The resulting of voxel latents $\{\tilde{\mathcal{V}}\}$ provide a structured and memory-efficient representation ready for downstream refinement. In practice, we find $K = \max(\lfloor 0.25N_k \rfloor, 1)$, \ie keeping up to 25\% Gaussians per voxel provides sufficient compression without inducing quality degradation. 

\subsection{Refinement via Test-Time Training (TTT)}
% \todo{Reviewers complained this part did not have sufficient details. Try to make it clearer based on the feedback} 
To enable iterative editing, our refinement module must adapt to new user inputs on the fly without overwriting the knowledge of the original asset. We achieve this using Test-Time Training (TTT)~\cite{sun2020test, sun2024learning, wang2025test, behrouz2024titans, behrouz2025s, karami2025lattice, zhang2025test}, which updates a small set of ``fast weights''~\cite{schlag2021linear} while keeping the main network backbone frozen. This allows our model to become state-aware, integrating new information from a user's edit before applying it to the 3D scene.

\vspace{1mm}\noindent\textbf{Preliminary: Test-Time Training.}
Test-Time Training adapts a model to a new test sample by defining a self-supervised task at inference time. Within a transformer layer, input tokens are projected into queries ($q$), keys ($k$), and values ($v$) by linear layers with ``slow weights''. TTT introduces a small, adaptable mapping network, $f_W$, governed by a set of ``fast weights'' $W$. For a given test sample, these fast weights are updated via gradient descent on a self-supervised loss:
\begin{equation}
\mathcal{L}_{\text{adapt}} = \mathbb{E}_{(k, v) \sim \text{TestSample}} \left[ \ell_2(f_W(k), v) \right],
\label{eq:ttt_loss}
\end{equation}
where the goal is for $f_W$ to learn to reconstruct the value from the key for pairs drawn from the test sample. Once adapted, the module processes queries from the same sample to produce the final output $o = f_W(q)$. This mechanism separates stable, general knowledge (slow weights) from instance-specific adaptation (fast weights).

\begin{figure}[t!]
  \centering
  \includegraphics[width=\textwidth]{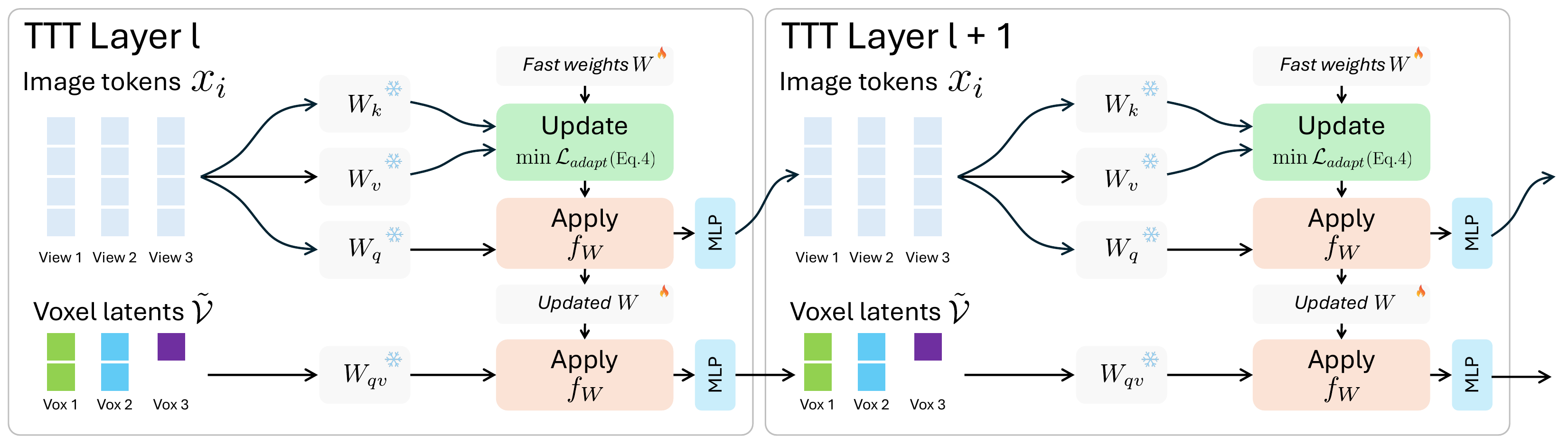}
  \caption{\textbf{TTT operations.} Fast weights $W$ are iteratively updated using new input views and subsequently applied to the voxel GS latents after seeing all the new inputs. During training, all the weights are optimized. During inference, the slow weights are fixed and only fast weights are updated. %The residual connections are omitted for clarity. \todo{redo this figure}
  }
  \vspace{-5pt}
  \label{fig:pipeline ttt} 
\end{figure}
\vspace{1mm}\noindent\textbf{Gaussian Refinement via TTT.}
Our refinement module is a deep transformer composed of a series of blocks, where each block contains a TTT layer followed by a feedforward network (MLP), as illustrated in Fig.~\ref{fig:pipeline}. Details of the network architecture are shown in \cref{fig:pipeline ttt}. Within each TTT layer, the model first adapts to the user's edit and then applies this adaptation in parallel to both the image and the voxel latents.

First, the fast weights $W$ of the mapping network $f_W$ are updated. The set of image tokens from all new user views, $X_{img} = \{x_i\}$, are projected into key and value vectors using frozen (slow) projection heads, $W_k$ and $W_v$:
\begin{equation}
k_i = W_k x_i, \quad v_i = W_v x_i, \quad \text{for } x_i \in X_{img}.
\end{equation}
The fast weights $W$ are updated by minimizing~\cref{eq:ttt_loss}) over all these $(k_i, v_i)$ pairs. This single update step infuses the TTT layer with the information from the entire user edit.

Once the fast weights are updated, the adapted mapping network $f_W$ is applied to two parallel processing streams.
For the image stream, the image tokens are projected into queries using a slow head $W_q$, and the adapted network produces updated image tokens:
\begin{equation}
x_{i} \leftarrow f_W(W_q x_i).
\end{equation}
For the 3D scene stream, the voxel latents $\tilde{\mathcal{V}}$ are projected into queries using a separate slow head $W_{qv}$, and the same adapted network produces the updated voxel latents:
\begin{equation}
\tilde{\mathcal{V}} \leftarrow f_W(W_{qv} \tilde{\mathcal{V}}).
\label{eq:refine}
\end{equation}
The outputs of both streams are then passed through their respective MLPs. The entire block is repeated multiple times, allowing the model to progressively refine the 3D asset by repeatedly conditioning on the user's input. Following recent work~\cite{zhang2025test}, we implement the parametric mapping $f_W$ as a SwiGLU-MLP~\cite{shazeer2020glu}.

\subsection{Applications}\label{sec:method-3} The unified design of our framework supports a wide range of editing tasks: \begin{itemize}[leftmargin=*] \item \textbf{Local Edits.} Our method enables refinement of high-frequency details from sparse close-up views (super-resolution) and integration of user-specified patches such as graffiti or text edits (see Fig.~\ref{fig:teaser}). These edits are applied to the visible voxels corresponding to the edited input views. When a new view is provided, we perform ray tracing to identify voxels that first intersect with the camera rays, ensuring updates are localized to intended region and the rest of the details are naturally preserved. In the local region, we further update the original GS tokens using refined tokens by adding an extra TTT layer, then merge the two token sets—effectively doubling the number of Gaussians to introduce additional fine-grained structures.
Please refer to the supplementary material for additional implementation details.
\item \textbf{Global Edits.} In addition to local refinement, our network also supports consistent global modifications. By applying the updated fast weights from TTT layers to all Gaussian tokens, our method enables coherent changes to appearance attributes across the entire scene. This design allows users to perform holistic edits while preserving geometry and fine structures of the original asset. We demonstrate relighting as a representative example. \end{itemize} 

Together, these components constitute \ourmethod: a feedforward, state-aware framework for interactive 3DGS refinement.  
Both local and global editing modules are trained with an image-based reconstruction loss:  
\begin{equation}
\mathcal{L}_{\text{recon}} = \| I_{\text{render}} - I_{\text{gt}} \|_2 + \lambda_{\text{perc}} \| \phi(I_{\text{render}}) - \phi(I_{\text{gt}}) \|_2, 
\label{eq:loss}
\end{equation}
where $I_{\text{render}} = R(\mathrm{MLP}(\tilde{\mathcal{V}}), P_v)$ denotes the rendered image from the refined Gaussians decoded from $\tilde{\mathcal{V}}$ (note here we decode all the GS attributes, i.e., positions, SHs, opacity, rotation such that both geometry and appearance are updated) under camera pose $P_v$, and $\phi(\cdot)$ is the LPIPS feature extractor~\cite{zhang2018perceptual}. 

\section{Experiment}
In this section, we explain our experimental setup, baselines, and evaluation results across our three primary applications: local refinement, creative editing, and global relighting.
\label{sec:exp}
\subsection{Setup}
\noindent\textbf{Data.} We use Objaverse1.0~\cite{deitke2023objaverse} and partial TexVerse~\cite{zhang2025texverse} for training. For local editing tasks, we render 32 2K-resolution images and simulate user inputs: zoomed-in crops for refinement, and overlay strokes/graffiti for creative editing. 
For relighting, we render 32 $512\times512$ images under 24 lighting conditions using randomly sampled HDRI maps. Evaluation is performed on 370 held-out assets from TexVerse, selected for their high-frequency texture details.
The evaluation data consists of 370 examples from TexVerse that are not used for training filtered by auto-captioning, focusing on assets with high-resolution textures and rich high-frequency details to better showcase our model’s capability in fine-grained refinement.

\vspace{1mm}\noindent\textbf{Baselines.} We compare against state-of-the-art methods tailored to each task. 
\begin{itemize}[leftmargin=*]
    \item \textit{For Local Refinement:} We compare against direct reconstruction ({GS-LRM}~\cite{gslrm2024}), feedforward densification ({GenDen}~\cite{nam2025generative}), and optimization approaches including 3DGS~\cite{kerbl3Dgaussians} and SRGS~\cite{feng2024srgs} specialized for 3DGS refinement.
    \item \textit{For Creative Editing:} We compare against {VoxHammer}~\cite{li2025voxhammer}, a recent method for editing via structured 3D latents. 
    \item \textit{For Global Relighting:} We compare against {ReLitLRM}~\cite{zhang2024relitlrm}, a generative relighting model and direct 3DGS optimization.
\end{itemize}
\textit{Note on Multiview Diffusion:} Ideally, we would compare against image-conditioned multiview diffusion methods (the ``Edit-Generate-Lift'' paradigm). However, current state-of-the-art methods in this category are either strictly text-conditioned~\cite{haque2023instruct, chen2024dge} or have not released code (e.g., CMD~\cite{li2025cmd}). To illustrate the limitations of this paradigm, we provide a qualitative comparison with {DGE}~\cite{chen2024dge} in \cref{fig:teaser2}, but we exclude it from quantitative tables due to the input modality mismatch.

\vspace{1mm}\noindent\textbf{Implementation Details.} 
In Stage I, the Gaussian LRM is based on GS-LRM~\cite{gslrm2024}, extended to produce view-dependent Gaussians with 4 orders of SH. The local transformer contains six encoder self-attention blocks, one cross-attention block, and six decoder self-attention blocks (see \cref{fig:pipeline}), each using 512 dimensions and 8 heads implemented with Flash-Attention~\cite{dao2023flashattention2}.
In Stage II, the refinement network has 8 TTT layers. The fast weights are obtained by taking 5 gradient descent steps using the Muon-update rule following Zhang et al.~\cite{zhang2025test}. The updated voxel latents $\tilde{\mathcal{V}}$ are decoded into Gaussian attributes through one layer MLP.
We first train the local transformer in stage I, then freeze it to train the refinement network. Both stages use image reconstruction loss (\cref{eq:loss}) with $\lambda_{perc}=0.5$. 
Stage I uses 9 to 16 input images; Stage 2 trains with 1 to 4 new inputs. All inputs are $512\times512$, downsampled from renderings, except for refinement views which are cropped from 2K images. Training uses 16 A100 80GB GPUs; testing is reported on a single A100.

\subsection{Evaluation on Local Editing}
\label{sec:exp:refinement}
We evaluate two distinct local editing scenarios: \textit{High-Fidelity Refinement}, where the goal is to recover missing details from zoomed-in views, and \textit{Interactive Creative Editing}, where the goal is to apply artistic changes (paint-overs, generative edits) while preserving the asset's identity.

\subsubsection{High-Fidelity Refinement}
In this task, a user supplies high-resolution 2D zoom-in views to refine specific regions of the 3D asset. We use 9 base views for initialization and 2 additional zoomed-in views for refinement.

\begin{figure}[t!]
  \centering
  \vspace{-10pt}
  \includegraphics[width=\textwidth]{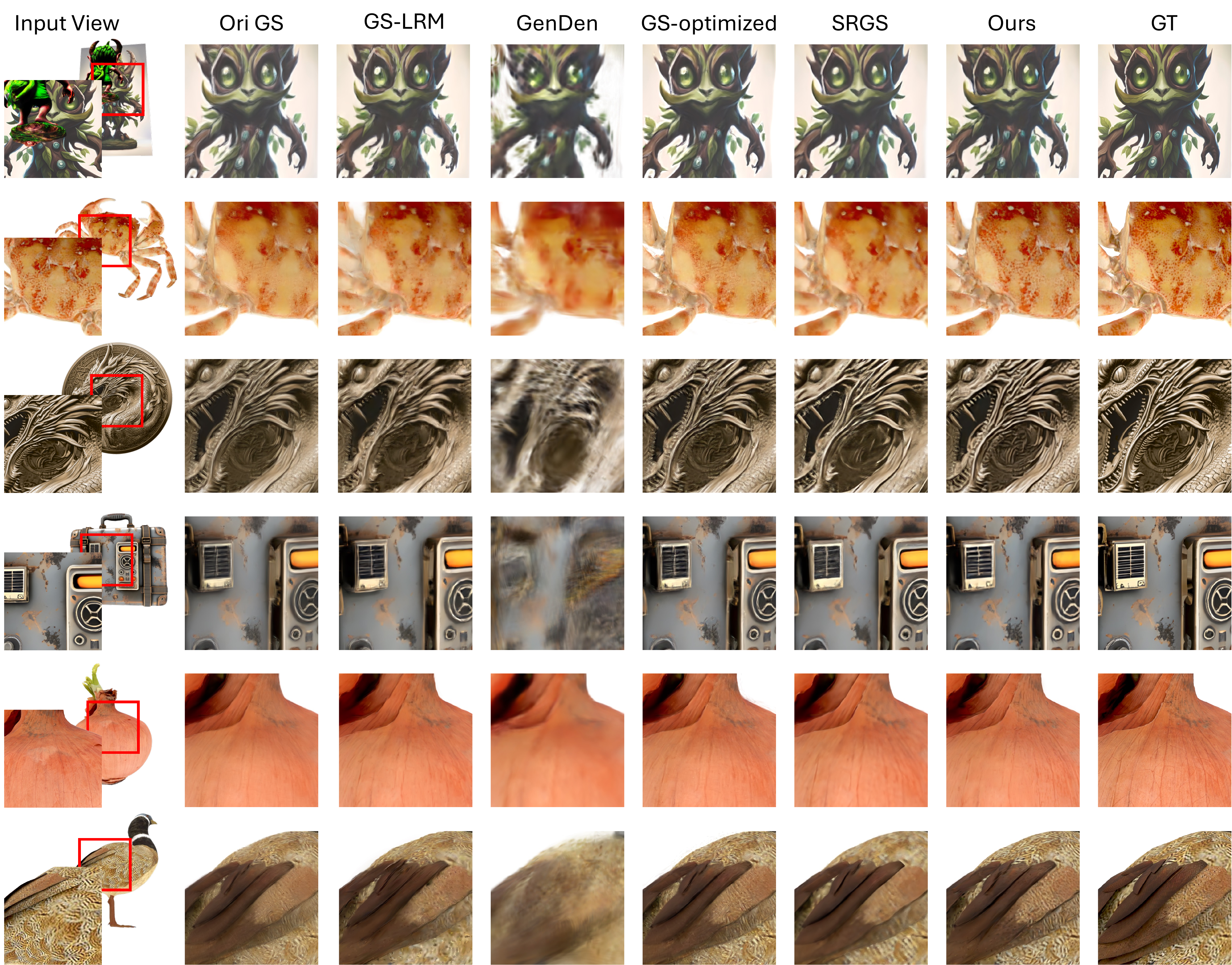}
  \caption{\textbf{Qualitative evaluation for local refinement}. Given a zoomed-in input view (left), we compare the refined results between baseline methods and ours from a novel view close to the zoomed-in view. Our method recovers the details provided by the input zoomed-in views, and produces much sharper features compared to direct reconstruction (GS-LRM~\cite{gslrm2024}) or upsampling method (GenDen~\cite{nam2025generative}), and is on-par or better than optimization-based approaches but using a fraction of the time (see \cref{tab:refinement}).
  % \yifan{We should order the columns same way as the table.}
  }
  \label{fig:refinement}
  \vspace{-10pt}
\end{figure}

\vspace{1mm}\noindent\textbf{Results.} We evaluate image quality on novel views sampled near the edit region. As shown in \cref{tab:refinement} and \cref{fig:refinement}, our method achieves the best visual and quantitative performance. 
Compared to feedforward methods (GS-LRM~\cite{gslrm2024}, GenDen~\cite{nam2025generative}), our method recovers significantly sharper high-frequency textures. While optimization-based methods (~\cite{kerbl3Dgaussians,feng2024srgs}) can achieve good sharpness, they require hundreds of seconds per update. Our method achieves superior visual quality at interactive speeds ($\sim$0.3s per edit after initial setup), enabling fluid online applications.

\begin{table}[t!]
    \centering
    % LEFT TABLE: Local Refinement
    \begin{minipage}[t]{0.48\textwidth}
        \centering
        \caption{\textbf{Quantitative evaluation for local refinement.} \ourmethod achieves the best performance on quality and efficiency, enabling interactive online refinement while preserving high-fidelity details. }
        \vspace{-10pt}
        \label{tab:refinement}
        \resizebox{\linewidth}{!}{
            \begin{tabular}{ccccc}
            \toprule
            Method & PSNR (\(\uparrow\))  & SSIM (\(\uparrow\)) & LPIPS (\(\downarrow\)) & Time (\(s\)) \\
            \midrule
            GS-LRM~\cite{gslrm2024} & 20.28 & 0.5185 & 0.3873 & 0.92\\
            GenDen~\cite{nam2025generative} &  17.32 &  0.5378 & 0.5979 & 1.68  \\
            \midrule
            3DGS~\cite{kerbl3Dgaussians} & 20.52 & 0.6553 & \textbf{0.3399} & 323\\
            SRGS~\cite{feng2024srgs} & 21.01 & \textbf{0.6759} & 0.3540 & 495\\
            \midrule
            Ours (\textit{stage I})& 20.84 & 0.5698 & 0.3782 & 1.54\\
            Ours (\textit{stage II}) & \textbf{21.07} & 0.5780 & 0.3673 & 1.54+0.64\\
            \bottomrule
            \end{tabular}
        }
    \end{minipage}
    \hfill % Adds flexible space between the two minipages
    % RIGHT TABLE: Global Relighting
    \begin{minipage}[t]{0.48\textwidth}
        \centering
            \caption{\textbf{Quantitative results for global relighting.} \ourmethod achieves the best reconstruction accuracy while maintaining interactive speed. Performance improves steadily as more relit views are provided, demonstrating robust global propagation of lighting changes.}
            \vspace{-10pt}
        \label{tab:recoloring}
        \resizebox{\linewidth}{!}{
            \begin{tabular}{ccccc}
            \toprule
            Method & PSNR (\(\uparrow\))  & SSIM (\(\uparrow\)) & LPIPS (\(\downarrow\)) & Time (\(s\))\\
            \midrule
            RelitLRM~\cite{zhang2024relitlrm} & 21.68 & 0.8726 & 0.0912 & 22.1 \\
            3DGS~\cite{kerbl3Dgaussians} &  21.32 & 0.8613 & 0.1225 & 320 \\
            GS-LRM~\cite{gslrm2024}+CrossAtt & 24.05 & 0.8970 & 0.0802 & 1.54+0.13\\
            \midrule
            Ours (\textit{1 view})& 24.56 & 0.8986 & 0.0721 & 1.54+0.08\\
            Ours (\textit{2 views}) & 24.71 & 0.8991 & 0.0719 &  1.54+0.11\\
            Ours (\textit{4 views}) & \textbf{25.03} & \textbf{0.9011} & \textbf{0.0712} & 1.54+0.20\\
            \bottomrule
            \end{tabular}
        }
    \end{minipage}
    \vspace{-5pt}
\end{table}

\subsubsection{Interactive Creative Editing}
This task evaluates the model's ability to incorporate artistic edits—such as graffiti, manual paint-overs, or generative modifications via tools like Nano Banana~\cite{nanobanana2026}—into the 3D asset. The core challenge here is \textit{Identity Preservation}: applying the edit without distorting the unedited geometry or appearance.

\begin{figure}[ht]
  \centering
  \includegraphics[width=\linewidth]{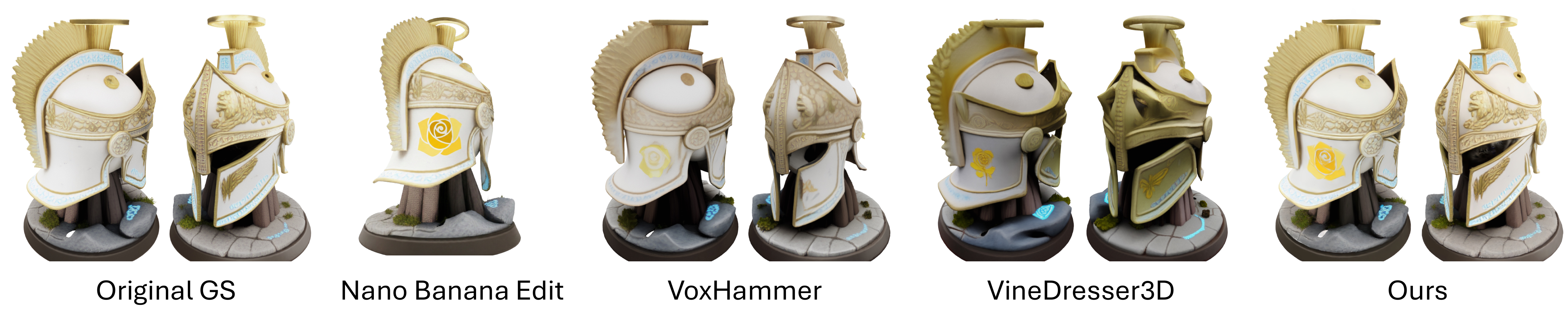}
  \caption{\textbf{Comparison of creative editing results}. 3DGS assets are generated from Trellis~\cite{xiang2024structured}. Here we show our results and VoxHammer~\cite{li2025voxhammer}, and VineDresser3D~\cite{chi2026vinedresser3d}.}
  \label{fig:edit}
  \vspace{-10pt}
\end{figure}

\begin{figure}[ht]
  \centering
  \includegraphics[width=\linewidth]{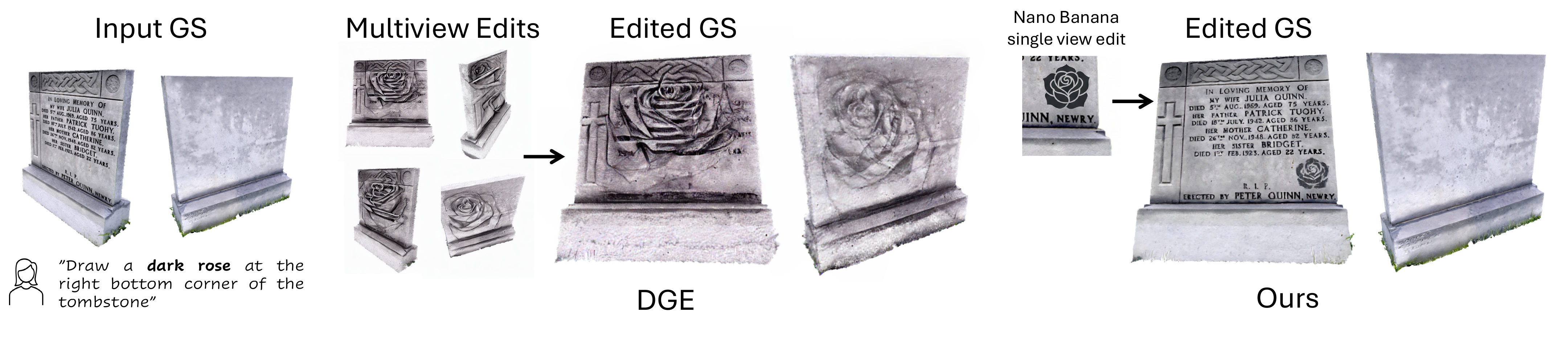}
  \vspace{-15pt}
  \caption{\textbf{Comparison of creative editing results} between DGE~\cite{chen2024dge} and ours.}
  \label{fig:edit_dge}
\end{figure}

\vspace{1mm}\noindent\textbf{Comparisons.} We compare our approach with VoxHammer~\cite{li2025voxhammer} and VineDresser3D~\cite{chi2026vinedresser3d}. For fair comparison, we use the same 2D edited image (generated by Nano Banana) as input for both methods. As illustrated in \cref{fig:teaser2} and \cref{fig:edit}, VoxHammer and VineDresser3D often struggle with ``spatial leakage'', where local edits cause unintended shifts in global geometry or color tone (e.g., the robot's eye shape or facial structure). In contrast, SplatPainter's voxel-based TTT update is performed locally, ensuring the edit is applied faithfully while the rest of the asset remains bit-exact to the original.
To quantify this, we measure the photometric losses of the edited and unedited regions of the assets (\cref{tab:identity_preservation}) on 10 generated examples from Trellis~\cite{xiang2024structured}. Our method achieves great identity preservation, confirming that our voxel-based updates prevent the identity drift common in latent diffusion approaches. For  multi-view generation methods, we choose a representative method, DGE~\cite{chen2024dge}, and show qualitative results in \cref{fig:edit_dge}, where DGE fails to interpret the fine-grained user editing intent and generates artifacts that undermine the original structure.

\begin{table}[t!]
    \centering

    % LEFT TABLE: Creative Editing
    \begin{minipage}[t]{0.48\textwidth}
        \centering
        \caption{\textbf{Quantitative evaluation for creative editing.}
        We measure photometric metrics on edited regions (compared to input edits) and non-edited regions before and after editing.}
        \vspace{-10pt}
        \label{tab:identity_preservation}
        \resizebox{\linewidth}{!}{
        \begin{tabular}{lccc @{\hspace{6pt}} ccc}
            \toprule
            \multirow{2}{*}{\textbf{Method}} &
            \multicolumn{3}{c}{\textbf{Edited Region}} &
            \multicolumn{3}{c}{\textbf{Non-edited Region}} \\
            \cmidrule(lr){2-4}\cmidrule(lr){5-7}
            & PSNR $\uparrow$ & SSIM $\uparrow$ & LPIPS $\downarrow$
            & PSNR $\uparrow$ & SSIM $\uparrow$ & LPIPS $\downarrow$ \\
            \midrule
            VoxHammer~\cite{li2025voxhammer} 
            & 11.25 & 0.2165 & 0.5036 
            & 13.86 & 0.4893 & 0.2906 \\
            VineDresser3D~\cite{chi2026vinedresser3d} 
            & 11.99 & 0.6092 & 0.3506 
            & 15.79 & 0.7437 & 0.2678 \\
            Ours 
            & \textbf{25.61} & \textbf{0.9380} & \textbf{0.0471}
            & \textbf{33.02} & \textbf{0.9560} & \textbf{0.0310} \\
            \bottomrule
        \end{tabular}}
    \end{minipage}
    \hfill
    % RIGHT TABLE: Voxel Compression
    \begin{minipage}[t]{0.48\textwidth}
        \centering
        \caption{\textbf{Evaluation of our voxel compression network.}
        Our LocalAtt + feats query design achieves the highest reconstruction accuracy with minimal quality loss.}
        \vspace{-10pt}
        \label{tab:ablation_vox}
        \resizebox{\linewidth}{!}{
        \begin{tabular}{lccc}
            \toprule
            Method & PSNR $\uparrow$ & SSIM $\uparrow$ & LPIPS $\downarrow$ \\
            \midrule
            VoxelAtt & 30.65 & 0.9193 & 0.0804 \\
            LocalAtt + latent query & 33.84 & 0.9349 & 0.0264 \\
            LocalAtt + feats query & \textbf{34.02} & \textbf{0.9387} & \textbf{0.0262} \\
            \midrule
            Pre-Compression (GS-LRM) & 34.19 & 0.9283 & 0.0249 \\
            \bottomrule
        \end{tabular}}
    \end{minipage}
\vspace{-10pt}
\end{table}


\begin{figure}[ht!]
  \centering
  % \vspace{-10pt}
  \includegraphics[width=\textwidth]{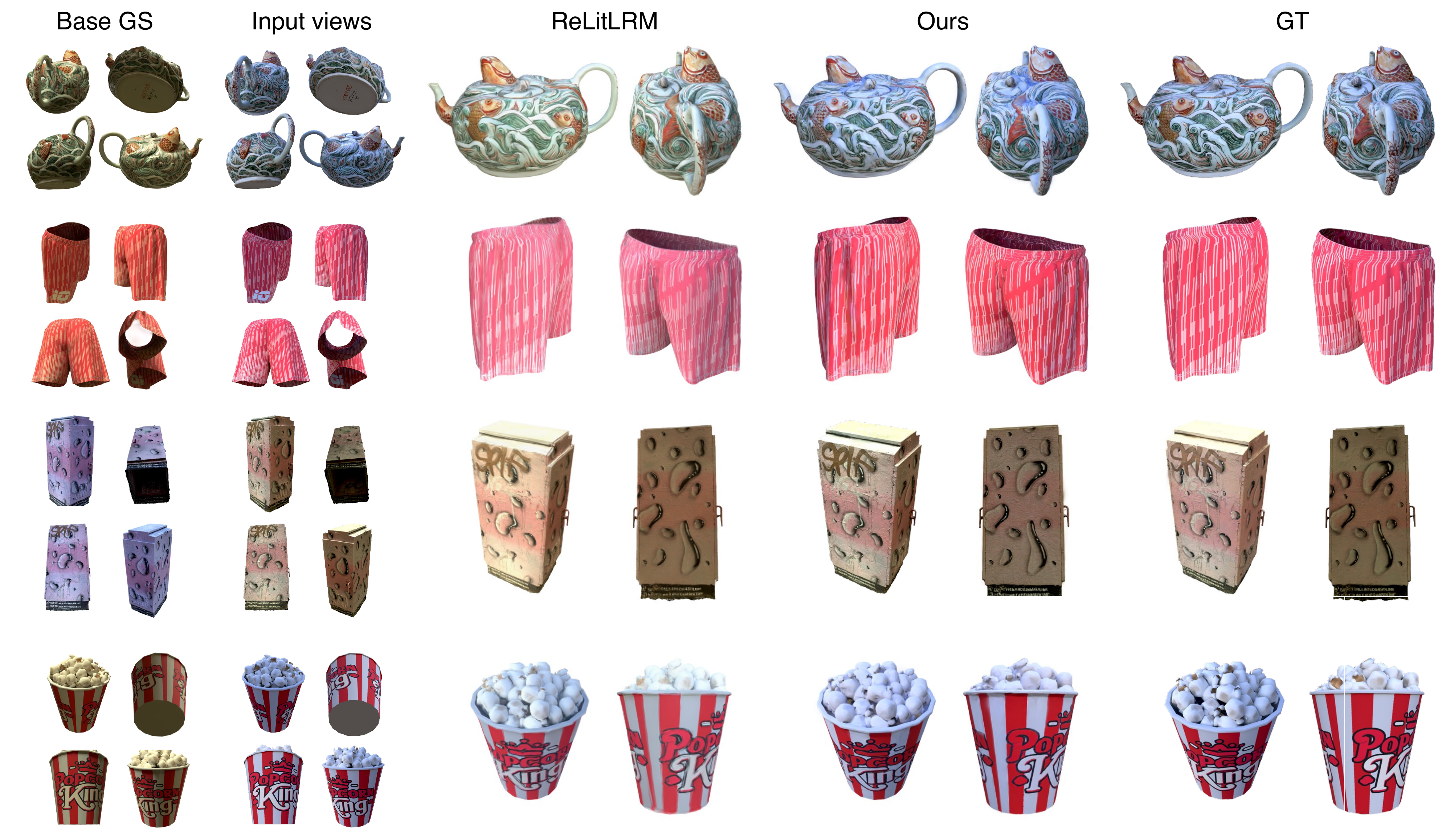}
  \caption{\textbf{Global relighting comparison}. 
  ReLitLRM takes an environment map as input to synthesize new lighting, while our method leverages a few relit views as direct input. Though not strictly relighting, our method offers a practical path for appearance transfer, achieving consistent shadows, accurate color propagation, and closer alignment with ground truth.
  % \yifan{Briefly say how input and outputs for RelitLRM and ours, and be humble, say that our approach is not exactly relighting and we receive more information than RelitLRM, this shows a viable way for relight GS.} Our method achieves more faithful recoloring with consistent shadows and accurate color transfer across views, closely matching the ground-truth.
  }
  \vspace{-10pt}
  \label{fig:recoloring}
\end{figure}

\subsection{Evaluation on Global Editing}
\label{sec:exp:recoloring}
Finally, we evaluate global refinement through relighting. The user provides one or more ``relit'' views (simulating a lighting change), and the model must propagate this new illumination globally.

\vspace{1mm}\noindent\textbf{Results.} \cref{tab:recoloring} compares our method against ReLitLRM~\cite{zhang2024relitlrm}. Our approach consistently achieves higher reconstruction accuracy. Notably, ReLitLRM often suffers from color tone mismatches (see \cref{fig:recoloring}), whereas our TTT-based propagation ensures consistent shading across views. Furthermore, our performance scales with user input: adding more views steadily improves fidelity (\cref{fig:ablation}), demonstrating the robust adaptability of our state-aware architecture.

\subsection{Analysis}

\begin{figure}[t!]
  \centering
  \includegraphics[width=\linewidth]{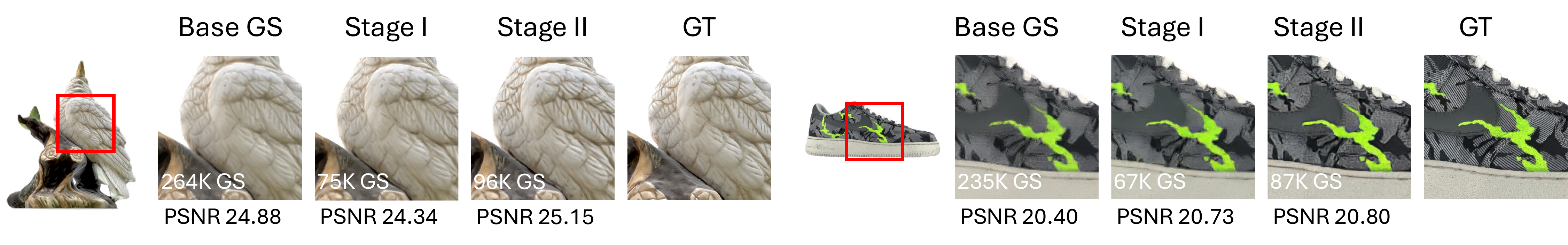}
  \caption{\textbf{The two-stage pipeline on the local refinement task.} Stage I achieves significant compression while maintaining visual fidelity. Stage II refines these compact GS, improving detail and overall quality based on the user's input.
  } 
  \label{fig:voxel_ablation}
\end{figure}

\vspace{1mm}\noindent\textbf{GS-Voxel Compression.} 
The goal of Stage I is to create a compact Gaussian representation while maintaining the visual fidelity necessary for editing. In \cref{tab:ablation_vox}, we quantify this by comparing different voxelization strategies against the uncompressed \textit{Pre-Compression (GS-LRM)} baseline.
Our chosen design (\textit{LocalAtt + feats query}) achieves the highest accuracy. Alternative designs like \textit{VoxelAtt} (averaging tokens) or \textit{LocalAtt + latent query} (generic query) result in significant information loss, as evidenced by the lower PSNR and SSIM scores.
To verify the effectiveness of our two-stage approach, \cref{fig:voxel_ablation} illustrates the visual quality and Gaussian count at each step. Stage I achieves aggressive compression—reducing the number of Gaussians by approximately while maintaining high visual fidelity. Stage II then builds upon this efficient representation, refining the high-frequency details to surpass the initial quality.

\begin{figure}[t!]
  \centering
  \includegraphics[width=\linewidth]{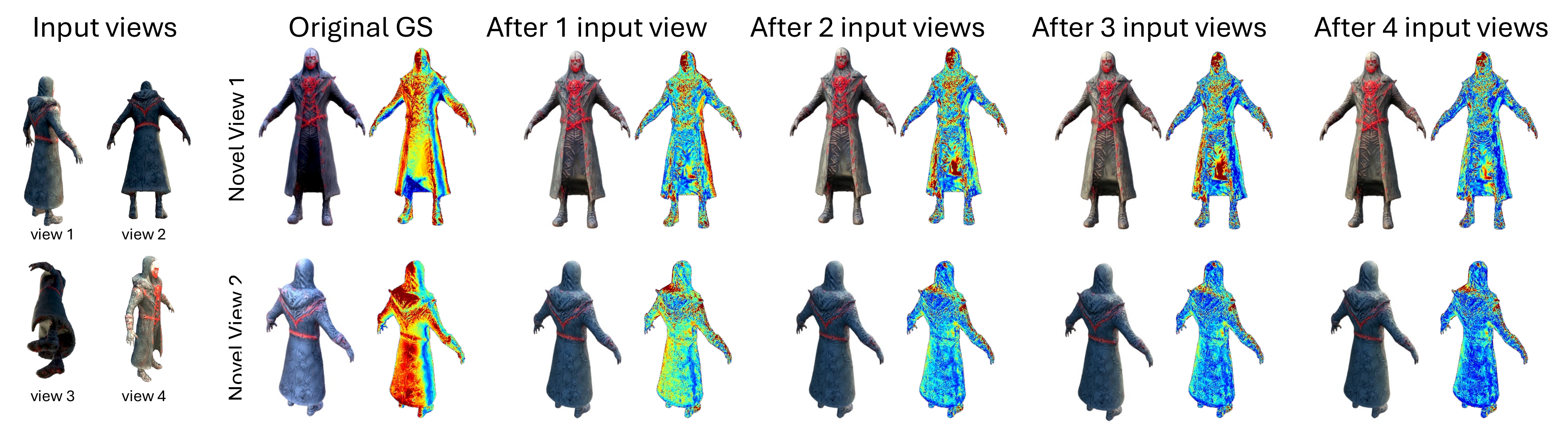}
  \caption{\textbf{Iterative updating on a global relighting task.} From left to right: original GS, refined GS with 1-4 input views, where we show the rendered images and error maps. Our model picks up partial shading hint (see the first input view) and propagates it coherently to the unseen part.
  } 
  \label{fig:ablation}
  \vspace{-10pt}
\end{figure}

\begin{figure}[t!]
  \centering
  \includegraphics[width=0.95\linewidth]{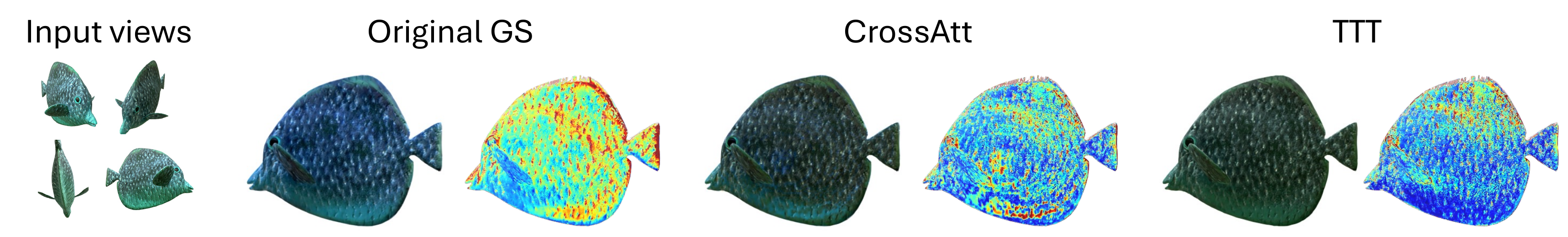}
  \caption{\textbf{Comparison between cross attention update and TTT update}. Our TTT design leads to more coherent relighting results.} 
  \vspace{-10pt}
  \label{fig:ablation_crossatt}
\end{figure}

\vspace{1mm}\noindent\textbf{TTT Update.}
As shown in \cref{fig:ablation}, our refinement quality improves with additional views. We compare our TTT refinement against a cross-attention baseline (replacing TTT layers with cross-attention). The variant achieves lower PSNR ( \cref{tab:recoloring}) and could introduce visual artifacts (\cref{fig:ablation_crossatt}). In contrast, our TTT design leads to more coherent global propagation.

\begin{figure}[ht!]
  \centering
  \includegraphics[width=\linewidth]{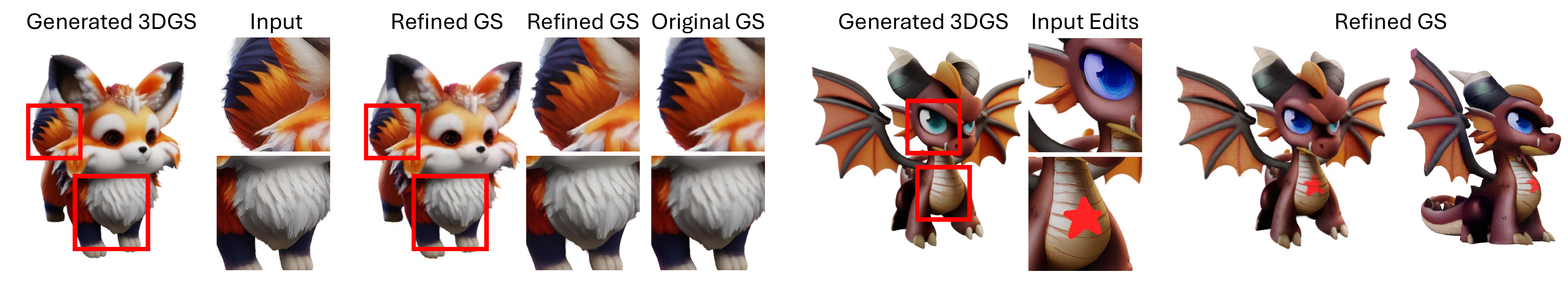}
  \vspace{-15pt}
  \caption{\textbf{More results}. We show our refinement results on generated 3DGS assets from Trellis~\cite{xiang2024structured} given local super-resolution and editing (graffiti, recoloring) inputs (better viewed zoomed in).}
  \label{fig:trellis}
  \vspace{-15pt}
\end{figure}

\vspace{1mm}\noindent\textbf{Generalization.}
We further test our model on generated 3DGS from Trellis~\cite{xiang2024structured}. As shown in \cref{fig:trellis} and \cref{fig:edit}, our method generalizes well to generated assets, refining geometry-aligned textures from both super-resolution inputs and creative edits.

\begin{figure}[ht]
  \centering  
  \vspace{-5pt}
  \includegraphics[width=\linewidth]{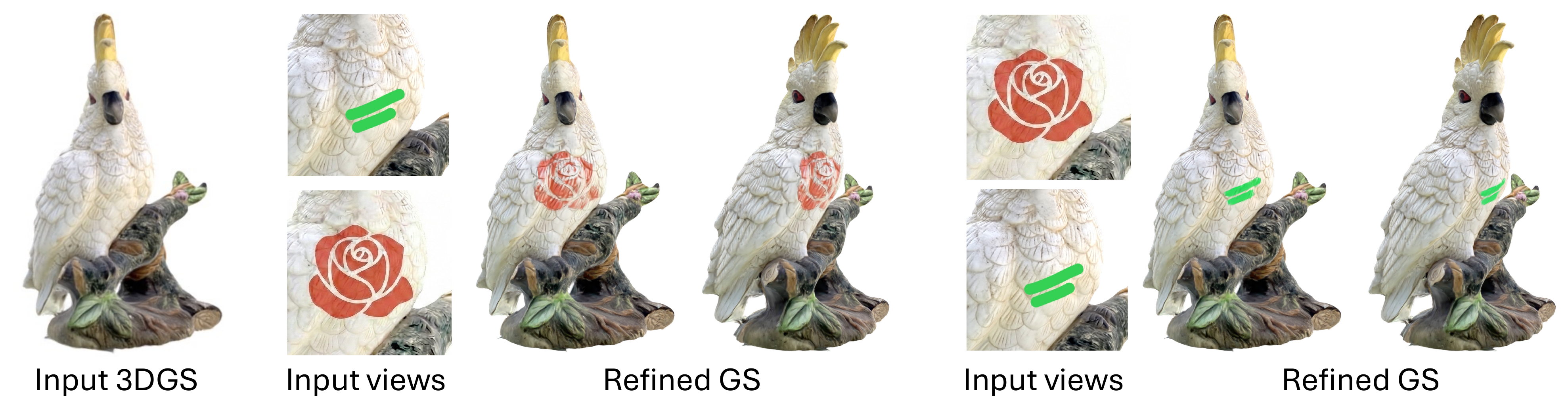}
  \vspace{-10pt}
  \caption{\textbf{Results of our method when handling conflict inputs}. Our model resolves conflict inputs by refining 3DGS based on the latest input information.}
  \label{fig:conflict}  
\end{figure}

\vspace{1mm}\noindent\textbf{Conflict Inputs.}
We further evaluate our method under conflicting edit inputs. As shown in \cref{fig:conflict}, different input views may introduce inconsistent edits to the same region. Our TTT refinement resolves these conflicts by updating the 3DGS representation according to the most recent input observation. As additional views arrive, the model progressively replaces outdated edits, producing a user desired final result.

\vspace{-10ex}\section{Conclusion and Future Work}
We presented \ourmethod, a feedforward, state-aware framework for interactive editing and refinement of existing 3D Gaussian Splatting (3DGS) assets.  
Our method introduces a compact voxel-based representation that substantially reduces redundancy while maintaining visual fidelity, and a test-time training (TTT) refinement module that efficiently incorporates new user inputs to achieve high-quality updates.  
Together, these designs enable both local and global editing tasks such as super-resolution, graffiti, and relighting, in an interactive manner.  

While our current design focuses on appearance updates and does not introduce new geometry, future work could extend the voxelization scheme to a multiresolution representation and incorporate geometric editing capabilities for more comprehensive scene manipulation. With suitable geometry-editing datasets, the proposed framework could be directly trained to support large-scale structural updates, which we leave for future exploration.

\clearpage
\appendix

% \maketitle
\begin{center}
\textbf{\Large Supplementary Materials}
\end{center}

\section{Implementation Details}
In this section, we provide additional details on the implementation of our method. 

\subsection{Network Implementation and Training}

\paragraph{Backbone and Feature Extraction.}
We adopt GS-LRM~\cite{gslrm2024} as our Gaussian large reconstruction model (LRM) backbone.
Given input multiview images (either coming from reconstruction inputs or rendered from a pre-generated asset), we first reconstruct a 3D Gaussian scene and its associated image tokens.
The image tokenizer and transformer backbone in GS-LRM are kept frozen in all of our experiments; we only use them to provide feature-rich tokens for our two-stage network.

\paragraph{Stage~I: Local Voxel Transformer.}
Stage~I compresses the dense GS-LRM output into compact voxel latents.
We voxelize the reconstructed 3DGS using voxel resolution as $128^3$ and group Gaussians into voxels as described in \textbf{Sec. 3.1}.
Each Gaussian is encoded by concatenating its GS-LRM feature token and Gaussian attributes (position, scale, opacity, rotation, spherical harmonics), followed by a linear projection to a $512$ dimensional embedding.

For each voxel, we apply a transformer encoder with $L_{\text{enc}}=6$ layers.
Each layer consists of multi-head self-attention (8 heads, hidden size 512) and a feedforward MLP with SwiGLU~\cite{shazeer2020glu} activation, both with residual connections and layer normalization.
We denote the encoded per-Gaussian latents as $\{z_k\}_{k=1}^{N_k}$.
To aggregate information, we select the top-$K$ Gaussians in each voxel based on opacity ($K = \max(\lfloor 0.25N_k \rfloor, 1)$) and use their latents as queries in a cross-attention layer.
A transformer decoder with $L_{\text{dec}}=6$ self-attention layers then produces $K$ compact latent per voxel, $\tilde{\mathcal{V}}$, which we store as our compressed GS representation.
Opacity-weighted queries focus on visible, perceptually dominant regions and empirically yield higher reconstruction quality (see \textbf{Tab. 3} in the main paper). For efficient batch training, all voxels are flattened and concatenated into a single 1D sequence, with per-voxel sequence lengths tracked explicitly.
We use FlashAttention2~\cite{dao2023flashattention2} in packed-attention mode to support variable-length voxel groups without padding overhead.

\paragraph{Stage~II: TTT Refinement Network.}
Stage~II refines voxel latents using new user-provided views.
The refinement module is a stack of $L_{\text{ttt}}=8$ TTT blocks (\textbf{Sec. 3.2}), each consisting of:
(i) a large-chunk TTT layer~\cite{zhang2025test} operating on image tokens and voxel latents, followed by
(ii) a feedforward MLP with SwiGLU~\cite{shazeer2020glu} activation.
The mapping network $f_W$ in each TTT layer is implemented as a 2-layer SwiGLU-MLP~\cite{shazeer2020glu} with hidden size 2048 and output dimension 512.
Fast weights $W$ are maintained per block, and only $W$ is updated during test-time training; all slow weights (projection heads, MLPs, and GS-LRM) remain frozen.

\paragraph{GS Decoder and Renderer.}
To map voxel latents back to Gaussians, we use a lightweight decoder consisting of one linear layer.
The decoder predicts Gaussian attributes (position, scale, opacity, rotation, SH coefficients).
We render the updated Gaussians using the original 3DGS renderer~\cite{kerbl3Dgaussians} to produce $I_{\text{render}}$.

% \paragraph{Training Data.}
% We train on Objaverse and TexVerse (4K and 8K subsets).
% For \emph{local refinement}, we render 32 views at $2048\times2048$ under a uniform HDRI, and randomly apply high-resolution graffiti and local texture perturbations on the underlying texture maps.
% We then downsample to $512\times512$ for training and take random $512\times512$ crops as the ``zoom-in'' views.
% For \emph{global relighting}, we render 32 views at $512\times512$ under 24 HDRI environment maps.
% During training, we randomly sample one lighting condition as the ``source'' GS and a different one as the ``target'' appearance, and supervise the model to reproduce the target lighting from the source GS plus a few target views.

% For evaluation, we handpick 370 TexVerse assets that are not used in training and exhibit high-resolution textures and strong high-frequency details, to better stress-test refinement quality.

\paragraph{Training Objectives.}
Both Stage~I and Stage~II are trained with an image-based reconstruction loss:
\begin{equation}
\mathcal{L}_{\text{recon}} = \| I_{\text{render}} - I_{\text{gt}} \|_2
\;+\;
\lambda_{\text{perc}} \big\| \phi(I_{\text{render}}) - \phi(I_{\text{gt}}) \big\|_2,
\end{equation}
as introduced in \textbf{Sec. 3.3}.
We set $\lambda_{\text{perc}} = 0.5$ in all experiments.
For Stage~I, $I_{\text{render}}$ uses randomly selected views.
For Stage~II, $I_{\text{render}}$ corresponds to: (i) zoomed-in crops for local refinement plus randomly selected views showing the whole objects, or (ii) globally relit views for the relighting task. 
We train Stage~I and Stage~II sequentially.
Stage~I is optimized for 80k iterations using AdamW with learning rate $4\times10^{-5}$, batch size 1, and cosine learning-rate decay.
We then freeze Stage~I (including the GS decoder) and train Stage~II for another 200k iterations with AdamW, learning rate $4\times10^{-5}$, batch size 1, and the same cosine schedule. Stage~I is trained using Objaverse~\cite{deitke2023objaverse} data, and we have two Stage~II models trained on different data based on  applications (local edits and global relighting). We aim at future work for training a unified model to support various applications.
All experiments are conducted on 16$\times$A100 80GB GPUs with mixed-precision training.

\subsection{Inference}

\paragraph{Preprocessing (One-Time per Asset).}
Given a new 3D Gaussian asset, we first run GS-LRM~\cite{gslrm2024} to obtain dense Gaussians and image tokens.
We then voxelize the scene and run Stage~I to obtain compact voxel latents $\tilde{\mathcal{V}}$ and the corresponding decoder that maps them back to Gaussians.
This preprocessing step is executed once per asset and takes $\sim$1.5\,s for the resolutions used in our experiments.

\paragraph{Interactive Editing Loop.}
At inference time, user edits are provided as a small number of 2D images:

\begin{itemize}[leftmargin=*]
\item \textbf{Local refinement.}
The user supplies one or more high-resolution zoom-in views, optionally with paint-over graffiti.
We associate each edited view with the visible voxels via ray tracing implemented based on Kaolin Library~\cite{KaolinLibrary} and only update latents in those voxels.
The image tokens of new input views are fed into Stage~II, where TTT layers update fast weights $W$ using these image tokens, and the updated $W$ is then applied to the corresponding selected visible voxel latents.
We decode updated Gaussians and re-render the scene.
A single local edit takes $\sim$0.3\,s (including ray-voxel association and rendering).

\item \textbf{Global relighting.}
The user provides one to four relit views under a new HDRI.
In this case, all voxel latents are refined by Stage~II without voxel-level masking or ray tracing.
The updated Gaussians are rendered under novel views to visualize the new lighting.
Each global update takes $\sim$0.06\,s, enabling interactive relighting.

\end{itemize}
\begin{figure}[t!]
  \centering
  \includegraphics[width=\textwidth]{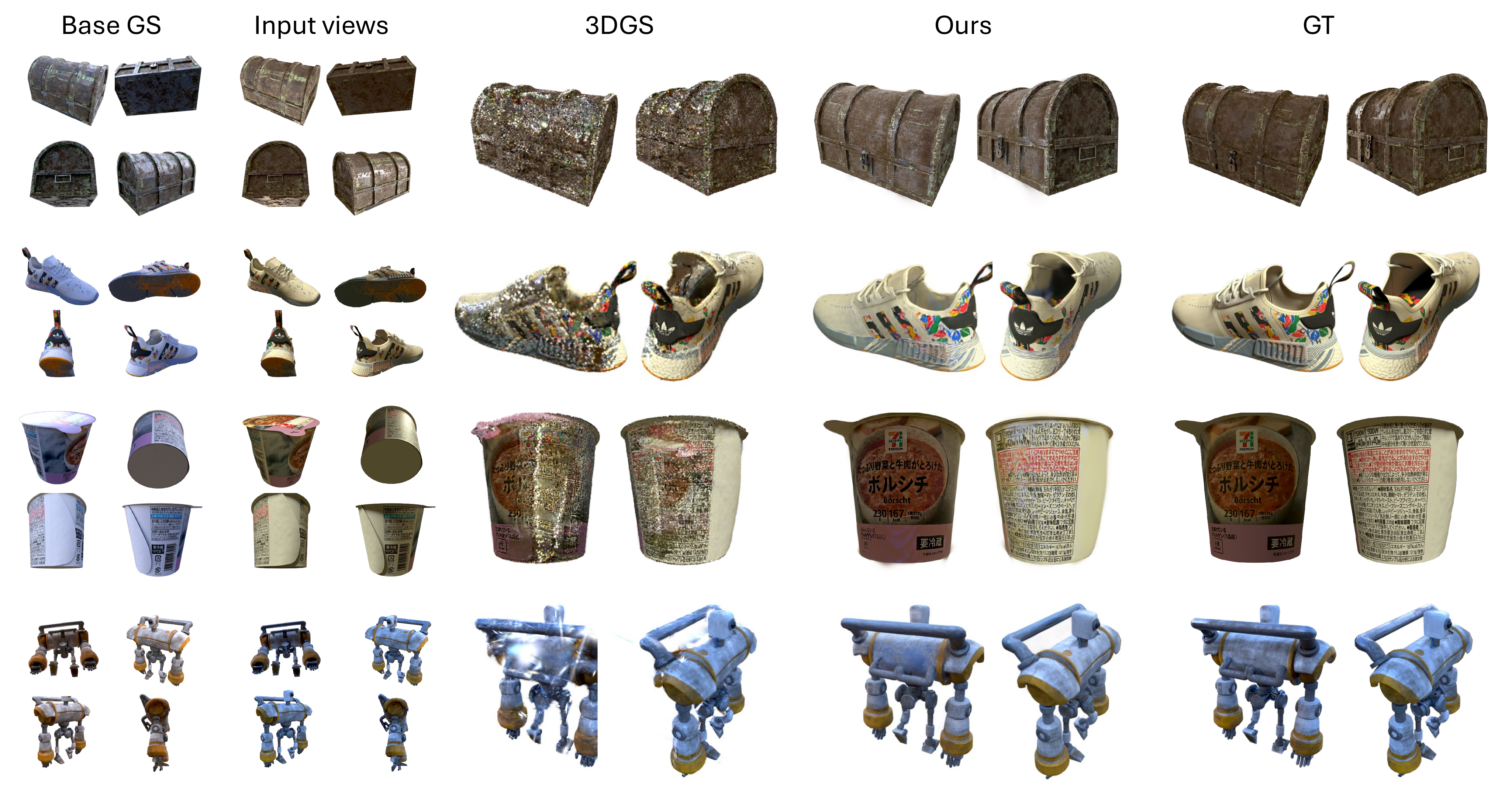}
  \caption{\textbf{Additional results of global editing.} Here we show our results and 3DGS optimization results on the global relighting task: our method generates coherent relighting results, while 3DGS optimization could lead to artifacts in the sparse view setting.}
  \label{fig:supp_relight}
\end{figure}

\subsection{Baseline Implementation}
We compare our method with state-of-the-art methods tailored to each task. We elaborate on implementation details below.

\begin{itemize}
    \item \textit{For Local Super-Resolution Refinement:} We compare our method with direct reconstruction method, i.e., GS-LRM~\cite{gslrm2024}, which takes inputs as all the 9 base views and 2 additional zoomed-in views and directly outputs the reconstructed 3DGS in a single forward run. Similarly, we also evaluate the feedforward densification method GenDen~\cite{nam2025generative} and optimization-based methods 3DGS~\cite{kerbl3Dgaussians} and SRGS~\cite{feng2024srgs} given all the 11 input views, where the input 3DGS is initialized from the output of the GS-LRM model using the 9 base views, and specifically for the optimization-based approaches, 48 additional uniformly sampled views of this initial 3DGS are provided during optimization to alleviate quality degradation under the sparse view optimization. All the baselines are evaluated based on the official implementations.

    \item \textit{For Creative Editing:} We compare against VoxHammer~\cite{li2025voxhammer} and VineDresser3D~\cite{chi2026vinedresser3d}, which take inputs as a generated 3D asset and an editing prompt from user input. The 3D asset is used to render multiview images, and then a best-fit single-view image (e.g., containing the region for editing) is chosen for image-based editing, given the input text prompt. The edited image is then fed to the Trellis~\cite{xiang2024structured} based pipelines to edit 3D geometry and appearance. For fair comparison, we use Nano Banana 2 to edit the image for the two baselines and our method. We also show qualitative comparisons between our method and a multi-view diffusion-based method, DGE~\cite{chen2024dge}, in the main paper (\textbf{Fig. 7}), where DGE takes text prompt input to perform multi-view consistent image editing.

    \item \textit{For Global Relighting:} we compare against ReLitLRM~\cite{zhang2024relitlrm} (a generative relighting model) and direct 3DGS optimization. For ReLitLRM, we use the original multi-view images and a GT target environment map (which is used to relit the original 3D asset to render the test views) as inputs.  To ensure fair comparison, we align ReLitLRM’s exposure levels with the GT relit views to mitigate lighting ambiguity. For 3DGS optimization, we initialize the GS as the output of GS-LRM given the original multi-view images, and optimize the SH and opacity parameters on the input relit views. 
\end{itemize}

\begin{figure}[t!]
  \centering
  \includegraphics[width=\textwidth]{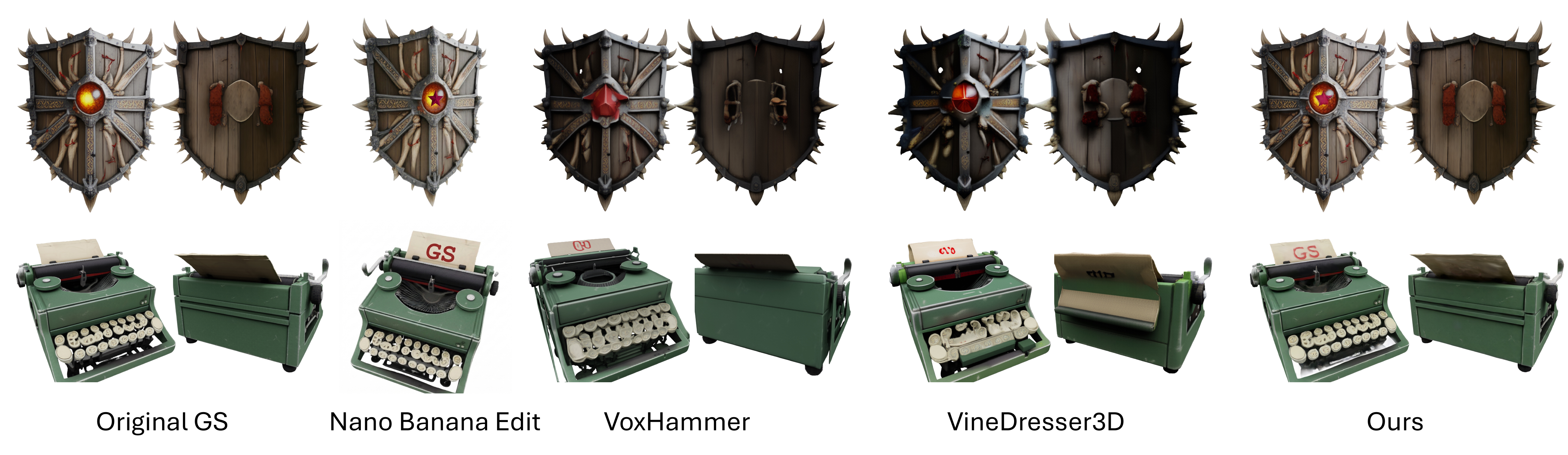}
  \caption{\textbf{Additional results of graffiti editing.} The input 3DGS assets are generated from Trellis~\cite{xiang2024structured}. Our method achieves precise editing while managing to preserve the original geometry structure and appearance. In contrast, baseline methods struggle to preserve the identity of the input assets.}
  \label{fig:supp_edit}
\end{figure}

\section{Additional Results}
\vspace{1mm}\noindent\textbf{Additional results of global editing.} We compare our method to 3DGS optimization in the global relighting task.
The 3DGS optimization pipeline takes input as the original 3DGS reconstructed by GS-LRM along with the input relit views, and during training, we fix the geometry attributes of GS and only optimize the appearance parameters. As shown in \cref{fig:supp_relight}, our method achieves robust relighting performance, while 3DGS optimization leads to visual artifacts in the sparse-view optimization setting, failing to propagate the new lighting information to the global appearance.

\vspace{1mm}\noindent\textbf{Additional results of creative editing.} We show additional results of graffiti editing in \cref{fig:supp_edit}. Our method achieves accurate graffiti editing results while maintaining the original geometric structure and appearance details. However, baseline methods fail to incorporate the detailed editing information into the original geometry and struggle to preserve the geometry and appearance identity.

Please refer to the website for additional visual results.

\bibliographystyle{splncs04}
\bibliography{main}
\end{document}